\pdfoutput=1
\documentclass[11pt]{article}

\usepackage{acl}

\usepackage{hyperref}
\usepackage{enumerate}
\usepackage{enumitem}
\usepackage{graphicx}
\usepackage{multirow}
\usepackage{booktabs}
\usepackage{makecell}

\usepackage{times}
\usepackage{latexsym}
\usepackage{amsmath}
\usepackage{amssymb}
\usepackage{mathtools}
\usepackage{amsthm}

\usepackage[T1]{fontenc}
\usepackage[utf8]{inputenc}
\usepackage{inconsolata}
\usepackage{microtype}
\usepackage{graphicx}
\usepackage{subcaption}
\usepackage{tcolorbox}
\tcbuselibrary{skins}

\colorlet{colexam}{blue!90!black}
\tcbset{
  base/.style={
    empty,
    frame engine=path,
    colframe=yellow!10,
    sharp corners,
    title={Example \thetcbcounter},
    attach boxed title to top left={yshift*=-\tcboxedtitleheight},
    boxed title style={size=minimal, top=4pt, left=4pt},
    coltitle=colexam,fonttitle=\bfseries\sffamily,
  }
}
\newtcolorbox[use counter=example]{myexamplea}{%
  base,
  boxed title style={overlay={
    \draw[colexam,line width=3pt,] (frame.north west)--(frame.north east);
  }},
  colback=colexam,
  overlay unbroken={
    \draw[colexam] ([yshift=-1.5pt]title.north east)--([xshift=-0.5pt, yshift=-1.5pt]title.north-|frame.east);
  },
}

\title{Weak Reward Model Transforms Generative Models into \\Robust Causal Event Extraction Systems}

\author{
	Italo Luis da Silva$^1$\quad Hanqi Yan$^1$\quad Lin Gui$^1$\quad Yulan He$^{1,2}$ \\
	$^1$King's College London \quad
    $^2$The Alan Turing Institute\\
	\texttt{\{italo.da\_silva,hanqi.yan,lin.1.gui,yulan.he\}@kcl.ac.uk}
}

\begin{document}

\maketitle

\begin{abstract}

The inherent ambiguity of cause and effect boundaries poses a challenge in
evaluating causal event extraction tasks. Traditional metrics like Exact Match
and BertScore poorly reflect model performance, so we trained evaluation models to
approximate human evaluation, achieving high agreement. We used them to perform
Reinforcement Learning with extraction models to align them with human
preference, prioritising semantic understanding. We successfully explored our
approach through multiple datasets, including transferring an evaluator trained
on one dataset to another as a way to decrease the reliance on human-annotated
data. In that vein, we also propose a weak-to-strong supervision method that
uses a fraction of the annotated data to train an evaluation model while still
achieving high performance in training an RL model.\footnote{Our code is
available at
\url{https://github.com/oyarsa/event_extraction/tree/causal-event-extraction}}

\end{abstract}
\section{Introduction}\label{introduction}

Causal event extraction is a crucial task in natural language understanding. It
involves identifying cause and effect clauses within an event and the
relationship between them. An example text along with its causal event
annotations from the Fine-grained Causal Reasoning (FCR)
dataset~\citep{yangFinegrainedCausalReasoning2022} is shown in
Figure~\ref{fig:fcr-example}. The emergence of powerful generative models leads
to a shift from span-based \textit{extraction} to the \textit{generation} of
structured
information~\citep{DBLP:journals/corr/abs-2311-02962,DBLP:journals/corr/abs-2310-03668}.
However, recent studies suggest that
ChatGPT~\citep{openaiGPT4TechnicalReport2023} struggles to surpass smaller
supervised models~\citep{DBLP:journals/corr/abs-2305-14450}, even when augmented
with Chain-of-Thought (CoT)~\citep{DBLP:conf/nips/Wei0SBIXCLZ22} and few-shot
In-Context Learning (ICL)~\citep{DBLP:conf/nips/BrownMRSKDNSSAA20}.

\begin{figure}[!ht]
    \centering
    \resizebox{\columnwidth}{!}{
    \begin{minipage}{1.25\columnwidth}
        \textbf{\texttt{Source Text}} 
        
        \textsl{\textcolor{teal}{The firm's gross margin is set to stabilize} as
		\textcolor{purple}{Harley refocuses its efforts on more profitable markets}, and
		our base case assumes that it stabilizes around 32\% in 2029, helped by a more
		measured approach to entering new markets.}

        \vspace{0.35em}
        \textbf{\texttt{Gold Extraction}} 
        
        \textcolor{purple}{Cause: \textsl{Harley refocuses its efforts on more profitable
		markets}} \\
        \textcolor{teal}{Effect: \textsl{The firm's gross margin is set to stabilize}} \\
        \textbf{Relation: \textsl{cause}}
    \end{minipage}
    }
    \caption{Example instance from the Fine-grained Causal Reasoning (FCR) dataset.}
    \label{fig:fcr-example}
\end{figure}

We focus on fine-tuning smaller language models using text annotated with causal
and effect spans for causal event extraction. However, we observe that unlike
traditional named entity recognition, where entities have clear and often
unambiguous boundaries, cause or effect spans may include intermittent text and
could have blurred word boundaries. This means that even with minor word
omissions, the semantic meaning of the cause and effect spans remains the same.
Consequently, the same text could have multiple valid annotations. Therefore,
training supervised models based on strictly matching only one set of valid
human annotations may result in less robust models. 



Evaluating causal event extraction is not straightforward. Evaluation metrics
based on direct token-level overlapping tend to neglect semantically valid
variations. Recent studies show that they do not align well with human
evaluations~\citep{DBLP:journals/corr/abs-2305-14450}. This issue could be
exaggerated under the generative settings~\citep{DBLP:conf/emnlp/QiZWZYLHLB23}.
While Large Language Models (LLMs) are considered an alternative in evaluating
the generation tasks due to their flexibility and ability to capture high-level
semantics, discrepancies still exist between GPT-3.5 evaluation outputs and
human evaluations, so human evaluators remain crucial to provide reliable
feedback~\citep{DBLP:journals/corr/abs-2101-00133}, despite the high cost.  


To address the high expense of human evaluation, we explore training evaluators
for causal event extraction to account for semantic variations. We sample event
extraction results from GPT-3.5 and a fine-tuned
FLAN-T5~\citep{chung2022scaling} model, inviting human annotators to
label the correctness of these extractions as `valid' or `invalid'. These human
evaluation results are then used to train an evaluator. Our experiments
demonstrate that an evaluator trained on a subset of human evaluations from one
dataset can be transferred to other datasets without losing alignment with the
actual human evaluation results.

Furthermore, we propose using the evaluator as a reward model to
fine-tune the causal event extraction model, FLAN-T5, through reinforcement
learning instead of traditional cross-entropy loss to prioritise semantic
similarity over exact matching. The Policy Proximal Optimisation
(PPO)~\citep{DBLP:journals/corr/SchulmanWDRK17} algorithm is used to align
generative models' behaviours with human preferences. In this method, a reward
model is first trained on human preference data and is used to produce feedback
scores, guiding the policy model to reinforce high scoring and penalise
low-scoring generations. 


In this paper, we incorporate the trained evaluator as the reward model into PPO
for causal event extraction. Our contributions are threefold:


\begin{itemize}[noitemsep,leftmargin=*,topsep=0pt]
    \item We built a causal relation extraction platform to collect human
    evaluation data, which is then used to train an evaluator (i.e. a reward
    model). It shows a 0.94 correlation with human evaluations.
    \item The reward model is integrated into the PPO algorithm for fine-tuning
    a FLAN-T5 model for causal event extraction. It achieves an average
    improvement of 4\% across three datasets. 
    \item To decrease the reliance on human evaluations and ground-truth references, we
    propose a weak-to-strong framework to fully exploit data efficiency of our
    proposed approach. We succeeded in using 50\% of the supervised data augmented
    by weak supervision with dynamic filtering as a reward model for RL
    training, obtaining comparable performance with the full reward model.
\end{itemize}

\section{Related Work}\label{sec:related-work}
We will introduce the recent work in causal extraction tasks, reward models
for reinforcement learning, weakly-supervised reward models and data
augmentation for generative models.

\subsection{Causal event extraction}
The goal of causal event extraction is to identify and extract cause and effect
events from an input text. Prior works focus on identifying relations between
entities, often trigger words~\citep{huguet-cabot-navigli-2021-rebel-relation,chen-etal-2020-joint,ma-etal-2022-joint}.
The works that focused on relations between events focus exclusively on simple
causal~\citep{mirza-tonelli-2016-catena,mariko-etal-2020-financial} relations,
with no fine-grained relations considered.

Existing works employed span-based extraction~\citep{becquinGBeFinCausal20202020a}
and sequence tagging~\citep{sahaSPOCKFinCausal20222022}, but they are limited to
single cause and effect scenarios, with simple relations. However, the recent increase in generative models, such as
T5~\citep{raffelExploringLimitsTransfer2020}, GPT-3.5 and
GPT-4~\citep{openaiGPT4TechnicalReport2023} highlight another possibility. They have shown the outstanding generalisation to not only learn
from IE training data through fine-tuning~\citep{paolini2021structured}, but
also extract information in few-shot and even zero-shot scenarios relying solely on in-context examples or
instructions~\citep{weiChainofThoughtPromptingElicits2022,wangSelfConsistencyImprovesChain2022}.
However, other works~\citep{nasar2021named,zhou2022learning} have shown
deficiencies in scenarios where there is a shortage of training data, an area
that has not been fully explored.
Traditional
metrics such as exact match (EM) and token F1 rely on the idea that a correct extraction is one that completely matches the annotation. 
There are other automated metrics such as ROUGE~\citep{linROUGEPackageAutomatic2004},
BLEU~\citep{papineniBLEUMethodAutomatic2001}, BLEURT~\citep{sellamBLEURTLearningRobust2020}
and BERTScore~\citep{zhangBERTScoreEvaluatingText2020} that attempt to
solve this problem, but we found them to not correlate well with
human annotations. Our solution was to train our own evaluation models
so that they correspond well with human evaluation. (Section~\ref{sec:evaluation}).

\subsection{Reward model in generative model}



Reinforcement Learning through Human Feedback
(PPO)~\citep{DBLP:conf/nips/Ouyang0JAWMZASR22} has seen applications for
instruction tuning~\citep{Shu2023RewriteLMAI,Lai2023OkapiIL}, controlled text
generation~\citep{Castricato2022RobustPL,Shulev2024ContinualRL},
summarisation~\citep{Roit2023FactuallyCS} and other generative
tasks~\citep{Cetina2021SurveyOR,Pang2023LanguageMS}. However, to the best of our
knowledge, it has not been applied to causal event extraction as a mechanism to
combat the limitations of automated metrics. Feedback acquisition is one of the
significant components, where humans or reward models assess the quality of the
base model’s responses to serve as a supervision signal for generative models. 


A critical aspect of this paradigm is to accurately model human preferences,
which involves the costly and time-consuming process of collecting feedback
data. Therefore, many recent works focus on how to fully steer the capabilities
of generative models with minimum
supervision~\citep{Yu2020FineTuningPL,Otani2022LITEIT}.

Several methods have improved LLMs by (self\nobreakdash-) creating training data to augment
fine-tuning. Self-Instruct~\citep{wang2022self} is a method for self-instruction
creation of prompts and responses, which can be used to improve a base LLM. 
Several approaches have also created training data by distilling
from powerful LLMs, and shown a weaker LLM can then perform well. For example,
Alpaca~\citep{taori2023alpaca} fine-tuned a Llama 7B model with text-davinci-003
instructions created in the style of self-instruct.
Alpagasus~\citep{chen2024alpagasus} employed a strong LLM-as-a-Judge (ChatGPT)
to curate the Alpaca dataset and filter to a smaller set, obtaining improved
results.

\section{Approximating Human Evaluation}\label{sec:evaluation}

Automated metrics for the evaluation of generated text have limitations in aligning with human evaluation. Metrics such as F1 score can measure the overlap
between the gold standard extraction and model outputs, but fail to recognise the
semantic aspects of such comparisons. In causal event extraction, we often
have situations where the output is different and has incomplete overlap with
the gold standard but is nonetheless correct. Automated metrics are unable to
deal with these situations since they cannot account for semantic differences,
such as when adding or removing words does not change the meaning of an
extraction.

One way to circumvent this issue is to employ human annotators to evaluate model outputs. While effective, it is
expensive and time-consuming, severely limiting experimentation and
the development of new approaches.

To address these limitations, we propose to collect human feedback to train an
evaluation model for high-quality feedback generation. The goal is to have an 
automated way to evaluate model outputs that approximates the judgement a human
would have made without the time-consuming and expensive aspects of human
evaluation.

\subsection{Human Feedback Collection}

\paragraph{Platform setup.} We built a platform to collect human annotations for
causal-effect extraction tasks. For each sample, annotators are given the
\textsl{Source Text}, \textsl{Cause} and \textsl{Effect}. For both
\textsl{Cause} and \textsl{Effect}, we provide the \textsl{Reference} and
\textsl{Model Output}. Annotators are asked to make a binary decision on
whether the \textsl{Model Output} is a valid extraction for the given source
text, with a sample only being valid if both \textsl{Cause} and \textsl{Effect}
are correct. See Section~\ref{sec:annotation-platform} (Appendix) for more
details.

To enhance the generalisability of the annotation data, we first apply
two different generative models, FLAN-T5 and GPT-3.5, to generate the cause and
effect results for evaluation. We remove
instances where the generated outputs are exact matches with the reference, as
those cases are trivial to evaluate. The remaining generated outputs are
organised using our tagged template. Figure~\ref{fig:fincausal-example} shows an
example instance from the FinCausal~\citep{mariko-etal-2020-financial} dataset,
including the \textsl{Source Text}, the \textsl{Cause} and \textsl{Effect}
spans, and the equivalent version in our tagged format.


\begin{figure}[!ht]
    \centering
    \resizebox{\columnwidth}{!}{
    \begin{minipage}{1.25\columnwidth}
        \textbf{\texttt{Source Text}}
        
        \textcolor{teal}{\textsl{It found that total U.S. health care spending would be about \$3.9 trillion under Medicare for All in 2019, compared with about \$3.8 trillion under the status quo}}. \textcolor{purple}{\textsl{Part of the reason is that Medicare for All would offer generous benefits with no copays and deductibles, except limited cost-sharing for certain medications}}.

        \vspace{0.35em}
        \textbf{\texttt{Gold Extraction}}

        \textcolor{purple}{Cause: \textsl{Part of the reason is that
        Medicare for All would offer generous benefits with no copays and
        deductibles, except limited cost-sharing for certain medications.}} \\
        \textcolor{teal}{Effect: \textsl{It found that total U.S. health care
        spending would be about \$3.9 trillion under Medicare for All in 2019,
        compared with about \$3.8 trillion under the status quo.}}

        \vspace{0.35em}
        \textbf{\texttt{Structured output (tagged format)}} 

        \textcolor{purple}{[Cause] \textsl{Part of the reason is that Medicare
        for All would offer generous benefits with no copays and deductibles,
        except limited cost-sharing for certain medications.}}
        \textbf{[Relation] \textsl{cause}} \textcolor{teal}{[Effect] \textsl{It found
        that total U.S. health care spending would be about \$3.9 trillion under
        Medicare for All in 2019, compared with about \$3.8 trillion under the
        status quo.}}
    \end{minipage}
    }
    \caption{An example instance from the FinCausal dataset. Top to Bottom: Source text in original dataset, Gold Standard Extraction, Structured output.}
    \label{fig:fincausal-example}
\end{figure}

The gold standard extractions for cause and effect are formatted into the same
structured output. Finally, both the formatted model output and the reference,
along with the \textsl{Source Text}, are presented to the annotators (shown in
the following \textbf{Examples}). Our instructions for annotators primarily
address the shortcomings of the current evaluation methods. We identify the two
most common issues: \textit{Wording Variation} and \textit{Hallucination}.

\paragraph{Pitfalls of Existing Evaluation Schema.} 
Two representative cases are
shown below. GPT-3.5 was used as an evaluator. In both cases, GPT-3.5's evaluation results differ from those of human evaluators. The evaluator errors for \textit{Wording Variation} always occur in the span
border, either adding some tokens or removing some tokens. The \textit{Hallucination} issue
happens when the generative model copies the text correctly but generates
incorrect numbers and symbols. These examples illustrate how
even a competent model struggles to reproduce human responses, motivating the
need for a specialised evaluation method.

\begin{myexamplea}\label{example1}
\scriptsize \textbf{\textcolor{blue!90!black}{Source Text}}: Our near-term earnings forecast is depressed due to the incorporation of crack spread futures curves despite a recent uptick.
\vspace{-4mm}
\tcbline
\textbf{\textcolor{blue!90!black}{Reference}}: [Cause] the incorporation of crack spread futures curves \textbf{despite a recent uptick} [Relation] cause [Effect] Our near-term earnings forecast is depressed.
\vspace{-4mm}
\tcbline
\textbf{\textcolor{blue!90!black}{Output}}: [Cause] the incorporation of poor crack spread futures curves [Relation] cause [Effect] Our near-term earnings forecast is depressed.
\tcbline
\vspace{-4mm}
\textbf{Evaluator}: Invalid \quad \textbf{Human:} Valid
\end{myexamplea}


\begin{myexamplea}
\scriptsize \textbf{\textcolor{blue!90!black}{Source Text}}: Analyst Ratings  This is a breakdown of recent ratings and recommmendations for Auris Medical and Elite Pharmaceuticals, as provided by MarketBeat.com.  Auris Medical currently has a consensus price target of \$75.00, indicating a potential upside of 2,383.44
\vspace{-4mm}
\tcbline
\textbf{\textcolor{blue!90!black}{Reference}}: [Cause] Auris Medical currently has a consensus price target of \$75.00 [Relation] cause [Effect] a potential upside of 2,383.44\%.
\vspace{-4mm}
\tcbline
\textbf{\textcolor{blue!90!black}{Output}}:[Cause] Auris Medical currently has a consensus price target of \$9.50 [Relation] cause [Effect] a potential upside of 655.21\%
\textbf{Evaluator}: Valid \quad \textbf{Human:} Invalid
\end{myexamplea}

\paragraph{Instructions For Human Annotators.} To alleviate the issues observed in the existing evaluation methods, we establish criteria for annotators. Only entries where both \textsl{Cause} and \textsl{Effect} satisfy all conditions should be considered as valid.

\begin{itemize}[noitemsep,topsep=0pt,leftmargin=*]
    \item Wording may differ between \textsl{Reference} and \textsl{Model Output}. This is fine, as long as the Model tokens come from the source text.
    \item There are no significant discrepancy between \textsl{Model Output} and \textsl{Reference}, such as numbers, subjects, time.
    \item If \textsl{Cause} and \textsl{Effect} happened to be in the same sentence but not overlapping, make sure the tokens in \textsl{Cause} are not included in the \textsl{Effect} and vice versa.
    \item In the rare cases where the \textsl{Reference} is obviously incorrect, ignore it and analyse the \textsl{Model Output} with relation to the source text only.
\end{itemize}

\subsection{Alignment with Human Feedback}

We conducted human evaluation on the extraction results from GPT-3.5 (10-shot) and FLAN-T5 on the training sets of the three datasets:
FCR~\citep{yangFinegrainedCausalReasoning2022},
FinCausal~\citep{mariko-etal-2020-financial} and SCITE~\citep{LI2021207}.\footnote{Dataset statistics are shown in Table~\ref{tab:datasets-count}.} The
Cohen's Kappa is 0.75, 0.51 and 0.84 for FCR, FinCausal and SCITE, respectively,
showing a good level of agreement between annotators on all datasets.

We use the extraction results from GPT-3.5 (10-shot) and FLAN-T5 to train evaluation models by obtaining 
human evaluations for the outputs of the training sets for FCR and FinCausal. These human-evaluated outputs were then used to train the evaluation models, while the development set outputs were used to evaluate their performance\footnote{We did not train
an evaluator on SCITE because the number of training samples is too small.},
with the guiding metric being agreement between evaluator outputs and the human
annotation~\citep{zhengJudgingLLMasaJudgeMTBench2023}. Our goal is for these
trained evaluators to approximate human judgement so we can use them as proxies for human evaluation in our experiments. 

Our evaluation model is the
DeBERTa-v3-based~\citep{heDeBERTaV3ImprovingDeBERTa2022} classifier, specifically the
\texttt{xsmall} variant, which we call DeBERTa-Valid. It takes both the source
text and the gold standard extraction as inputs, along with the model output, to
produce a classification. It is a binary classifier, with the positive class
referring to `valid' examples and the negative class to `invalid'. 
We also explore variations of the DeBERTa classifier: 

\begin{itemize}[leftmargin=10pt,itemsep=-3pt,topsep=2pt]

    \item \underline{DeBERTa-Entailment}: an instance is considered correct if
    there is an entailment between the extracted output and the original source
    text. Its inferior performance shows its inefficiency in evaluating the
    generated cause/effects.
    
    \item \underline{DeBERTa-Valid variants}: one variant excludes the reference
    extraction, and another excludes the source text. The poor performance of the
    variant without the reference shows its importance to our evaluator.
    Notably, the version without the source text also shows decreased agreement,
    indicating that the evaluator still needs it, as the references are not
    always reliable.
    
\end{itemize}

In addition, we use GPT-3.5 with or without self-consistency as additional
automated evaluators for the causal event extraction task. To verify the
effectiveness of our trained evaluator models, we calculate the
agreement between our evaluator outputs and human evaluations on the development set, along with categorical metrics such as Exact Match in Table~\ref{tab:metric-agreement}. We also examine the correlation between continuous metrics commonly used to evaluate extraction results, such as F1 and BertScore, and human evaluations. Pearson
correlation results are shown in
Tables~\ref{tab:metric-correlation}. In both tables, we observe the low scores of existing automated metrics, highlighting
their inability to replicate human evaluations.  In contrast, our trained DeBERTa-based
model achives higher agreement and correlation scores. 

The results lead to the following observations: (a) automatic metrics do not
align well with human evaluation. (b) LLMs demonstrate similar results to
SentenceTransformer~\citep{reimers-2019-sentence-bert} (SentTF), even with
advanced prompting techniques, such as CoT and
Self-Consistency~\citep{wangSelfConsistencyImprovesChain2022}. (c) Supervised
classification models (DeBERTa\nobreakdash-*) perform the best. The inclusion of the reference is particularly crucial, which allows the reward model to
achieve near-complete agreement with human evaluation.

We use DeBERTa-Valid, the best-performing model, as our proxy for human evaluation and the primary reward model in the following sections.



\begin{table}[ht]
    \centering
    \resizebox{0.48\textwidth}{!}{%
    \begin{tabular}{lcc}
        \toprule[1pt]
        \textbf{Metric}                 & \textbf{T5}          & \textbf{\makecell{GPT-3.5 \\ (10-shot)}}  \\
        \midrule

            Exact Match            & 55.60       & 72.04           \\

        \midrule
            GPT-3.5        & 64.85       & 35.88      \\
            GPT-3.5-self-consistency  & 85.58       & 77.92 \\
        \midrule

            DeBERTa-entailment    & 68.61      & 43.19       \\
            DeBERTa-Valid-w/o-Reference  & 65.03      & 35.98 \\
            DeBERTa-Valid-w/o-SourceText & 92.51      & 82.47 \\
            DeBERTa-Valid     & \textbf{94.08}      & \textbf{86.26} \\

        \bottomrule[1pt]
    \end{tabular}
    }

    \caption{Agreement between human annotations and different metrics/evaluators on FCR (continuous metrics omitted). Various metrics are used to evaluate causal event extraction results from T5 and GPT-3.5 (10-shot)}
    \label{tab:metric-agreement}
\end{table}
\begin{table}[ht]
    \centering
    \resizebox{0.45\textwidth}{!}{%
    \begin{tabular}{lcc}
        \toprule[1pt]
        \textbf{Metric}                 & \textbf{T5}          & \textbf{GPT-3.5 (10-shot)}  \\
        \midrule

            ROUGE-L                & 80.94       & 67.15              \\
            BLEU                   & 76.73       & 66.46      \\
            BLEURT                 & 77.93       & 68.63         \\
            BertScore              & 75.94       & 65.83    \\
            F1                     & 80.61       & 65.64 \\
        \midrule

            SentTF                    & 63.70       & 47.53  \\
        \midrule

            DeBERTa-Valid     & \textbf{87.04}      & \textbf{72.98} \\

        \bottomrule[1pt]
    \end{tabular}
    }

    \caption{Pearson correlation between human evaluations and different
    metrics/evaluators on FCR.}
    \label{tab:metric-correlation}
\end{table}



\paragraph{Transfer to other datasets.} While using human evaluation to
train an evaluation model leads to high-performing evaluators, this approach can
be costly, especially for large datasets. We propose an alternative: train an
evaluation model in one dataset and transfer it to others with similar structure.
This is supported by the agreement between different combinations of
evaluators and datasets, as shown in Table~\ref{tab:transfer-agreeement}. We
observe high agreement between the FCR evaluator and the transferred datasets'
human evaluations, demonstrating the evaluator's transferability. As a
result, we use the FCR evaluator as the default reward model and the evaluator for causal event extraction in our experiments.


\begin{table}[htb]
    \centering
    \resizebox{0.8\columnwidth}{!}{%
    \begin{tabular}{cccc}
    \toprule[1pt]
       \multirow{2}{*}{\textbf{Source $i \rightarrow$}}  & \multicolumn{3}{c}{\textbf{Target $j$}}  \\
       \cmidrule{2-4}
       &            FCR    & FinCausal & SCITE \\
       \midrule
       FCR        & \textbf{94.08}  & \textbf{92.04}     & \textbf{96.86} \\
       FinCausal  & 73.57  & 91.58     & 88.48 \\
    \bottomrule[1pt]
    \end{tabular}
    }
    \caption{Agreement between the feedback generated by the reward model
    trained on dataset $i$ and human evaluation, when applying this model to
    generate feedback for dataset $j$\protect\footnotemark.}
    \label{tab:transfer-agreeement}
\end{table}

\footnotetext{Because SCITE is a small dataset, we could not train an effective evaluator with it.
See Table~\ref{tab:datasets-count} for dataset statistics.}

\section{Causal Event Extraction with Weak Reward Model}
In this section, we introduce our Reinforcement Learning (RL) framework designed to align our
generative extractor with human preferences. We also describe our process for training a weakly supervised reward model, which aims to minimise the data
needed for train the reward model.

\subsection{Reinforcement Learning for Cause Event Extraction}





\begin{figure}[ht]
    \centering
    \includegraphics[width=0.48\textwidth]{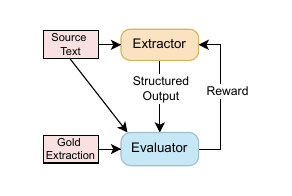}
    \caption{Architecture of our RL framework. PPO is used to optimise the extractor given the reward from the evaluator.}
    \label{fig:rl-arch}
\end{figure}

Our goal is to leverage the feedback from the trained evaluator described in
Section~\ref{sec:evaluation} to improve the generative extractor to be better
aligned with human preferences. See Figure~\ref{fig:rl-arch} for an overview of
our method.

We initialise an RL policy from the FLAN-T5 supervised fine-tuned extractor (our
reference model). It takes as input the source text and generates a structured
output representing the cause and effect using our tagged format
(Figure~\ref{fig:fincausal-example}). Both input and output are sequences of
tokens from the model vocabulary, which represents the action space. The policy
itself is a probability distribution over the action space conditioned on the
input tokens from the source text.

The RL objective is to find the optimal policy that maximises the reward. Our
reward is generated by the evaluation model described in
Section~\ref{sec:evaluation}. It takes as input the source text, the gold standard
extraction and the output from the RL policy, generating a scalar score. This is
done at the sequence level, as a complete extraction is needed to determine the
validity of the policy's output. Therefore, the score indicates whether the RL-generated extraction is valid, relative to the source text and the gold standard.

In addition to the reward model, we calculate the Kullback-Leibler (KL)
divergence to measure the disparity between our policy and reference models.
This helps us regulate the policy's ability to maintain the structured output format and prevent it from forgetting how to extract causes and effects. 
The final loss is a combination of the reward score and the KL divergence. We use
the Proximal Policy Optimisation (PPO) algorithm to update the policy parameters by
optimising this loss. During training, only the policy parameters are updated, while the reference and reward
models are frozen.

\subsection{Training a Weak Reward Model using Semi-Supervised Learning}\label{sec:weak-to-strong}

Our approach works well but relies on the performance of the reward model. While we have
trained a robust reward model, we explored scenarios with more limited data. 
To investigate this, we designed a weak-to-strong supervision process where we used a
small portion of our dataset to train the evaluator, treating the remaining data as 
unlabelled for further improvements.

We randomly sampled $x\%$ of our labelled training data, where $x$ is a hyperparameter. We
first trained a DeBERTa classifier 
reward model  
on the $x\%$ data. We then used this  classifier to generate
labels for the remaining data. To gauge the model's confidence in each example, 
we applied softmax to its outputs and retained only those examples where the predicted class probability ranked in the top 75\% separately for each class ('valid' and 'invalid'). This ensured an equal proportion of 'valid' and 'invalid' weak labels. 
Next, we
combined these filtered examples with the original partial dataset to create the
final weakly-supervised dataset, and trained a new DeBERTa model using this dataset.

Once we obtained a weakly-supervised reward model, we integrated it into our RL
process to develop an RL-trained model. We then compared the performance of
this new model with the original RL model trained with the full reward model.
We find that the weakly-supervised RL model has competitive performance with
the original RL model, demonstrating the effectiveness of our method
(Section~\ref{sec:weak_evaluator}).
\section{Experiments}\label{sec:experiments}

\paragraph{Datasets.} We employ three causal extraction datasets: FCR, FinCausal
and SCITE. Table~\ref{tab:datasets-count} shows statistics about them regarding
the number of examples in each split. Figure~\ref{fig:fincausal-example} shows
an example. Each entry contains an input context, cause and effect spans. These are
converted to our tagged format, which represents the relations textually.
Table~\ref{tab:datasets-words} (Appendix) shows more information.

\begin{table}[htb]
    \centering
    \resizebox{0,8\columnwidth}{!}{%
    \begin{tabular}{lccc}
    \toprule[1pt]
       \textbf{Dataset} & \multicolumn{3}{c}{\textbf{Number of examples}} \\
       & Train & Dev & Test \\
       \midrule
       FCR & 19892 & 2482 & 2433 \\
       FinCausal & 3397 & 641 & 817 \\
       SCITE & 1078 & 191 & - \\
    \bottomrule[1pt]
    \end{tabular}
    }
    \caption{Dataset statistics.}
    \label{tab:datasets-count}
\end{table}

\paragraph{Inplementation and Metrics.} We use FLAN-T5-Large as our policy model
and DeBERTa-v3-xsmall trained on human annotation data as our reward model
(Section~\ref{sec:evaluation}). For evaluation, we obtained the formatted
outputs from FLAN-T5-Large and gave them to our Human Proximal evaluator,
denoted as Human Prox.\footnote{This is the DeBERTa-Valid model trained with FCR
defined in Section~\ref{sec:evaluation}.}, along with the references and source
text. We also include automatic metrics such as Exact Match, Precision, Recall
and F1 for comparison.


\paragraph{Baselines.} We compare with another extractive IE model,
\textsl{Seq-tagging}, which is a sequence labelling model to predict
cause/effect BIO labels for each token. For the generative IE models, we compare
with our backbone model \textsl{FLAN-T5-Large}. We also compare with the
commercial large language models \textsl{GPT-3.5} and \textsl{GPT-4}, both
prompted with a structure generative format, using in-context learning. We also
report metrics from the original dataset
papers~\citep{yangFinegrainedCausalReasoning2022,mariko-etal-2020-financial,LI2021207}.


\subsection{Main results}


Table~\ref{tab:main_results} shows the causal relation extraction results of
various models across three datasets. We see that GPT-3.5 and GPT-4 underperform,
along with the other baselines, such as sequence tagging.

Our models perform much better, with the RL variant
achieving an improvement over the SFT version. This includes both automated
metrics and our
Human Proximal (Human Prox.) evaluator.

Our Human Proximal evaluator is the trained metric described in
Section~\ref{sec:evaluation}, which approximates the human preference. We show
that our supervised models achieve big improvement over both baselines and GPT
models, with the RL models further improving on them. As this happens on all
three datasets, we establish the superiority of our approach over the baselines.

\begin{table}[ht]
    \centering
    \resizebox{\columnwidth}{!}{
    \begin{tabular}{lrrrrr}
        \toprule
        & \textbf{P} &  \textbf{R} & \textbf{F1} & \textbf{EM} & \textbf{\makecell[c]{Human\\Prox.}} \\
        \midrule
        \multicolumn{6}{c}{\textbf{FCR}} \\
        \midrule

        \textsl{GPT-3.5} & 74.07 & 70.23 & 67.64 & 33.99 & 47.02  \\
        \textsl{GPT-4} & 74.53 & 69.27 & 64.70 & 28.24 & 39.66 \\
        \textsl{FCR-Baseline} 
            & - & - & 74.54 & 23.01 & - \\
        \textsl{Seq-tagging} & 77.76 & 77.78 & 77.74 & 41.30 & 52.82 \\
        \textsl{FLAN-T5-Large (SFT)} & 80.02 & 80.48 & 80.96 & 54.13 & 64.42 \\
        \textsl{FLAN-T5-Large (RL)} & \textbf{82.85} & \textbf{82.03} & \textbf{81.23} & \textbf{55.58} & \textbf{68.29} \\

        \midrule
        \multicolumn{6}{c}{\textbf{FinCausal}} \\
        \midrule

        \textsl{GPT-3.5} & 57.76 & 56.11 & 61.58 & 17.32 & 52.73 \\
        \textsl{GPT-4} & 63.35 & 61.92 & 66.58 & 26.99 & 55.85 \\
        \textsl{FinCausal-Baseline} & 50.99 & 51.74 & 51.06 & 11.11 & - \\
        \textsl{Seq-tagging} & 21.59 & 27.05 & 60.82 & 01.56 & 05.62 \\
        \textsl{FLAN-T5-Large (SFT)} & 78.19 & 77.93 & 78.52 & \textbf{66.61} & 81.12 \\
        \textsl{FLAN-T5-Large (RL)} & \textbf{88.60} & \textbf{88.70} & \textbf{88.64} & 64.74 & \textbf{84.40} \\

        \midrule
        \multicolumn{6}{c}{\textbf{SCITE}} \\
        \midrule

        \textsl{GPT-3.5} & 46.66 & 86.08 & 60.48 & 53.66 & 52.88 \\
        \textsl{GPT-4} & 37.97 & 83.70 & 52.23 & 46.86 & 57.59 \\
        \textsl{SCITE-Baseline}
            & 83.33 & 85.81 & 84.55 & - & - \\
        \textsl{Seq-tagging} & 92.94 & 92.25 & 92.59 & 88.48 & 91.10 \\
        \textsl{FLAN-T5-Large (SFT)} & 92.29 & 91.73 & 92.01 & 87.43 & 90.58 \\
        \textsl{FLAN-T5-Large (RL)} & \textbf{94.54} & \textbf{93.70} & \textbf{94.12} & \textbf{93.98} & \textbf{92.67} \\

        \bottomrule
    \end{tabular}
    }
    \caption{Causal relation extraction results on three datasets with automatic
    metrics and human evaluation showing our RL method performs the best in all
    three datasets.}
    \label{tab:main_results}
\end{table}

\subsection{Ablation results of our Reward model} 

\begin{figure}
    \centering

    \begin{subfigure}[h]{0.48\textwidth}
        \includegraphics[width=0.88\textwidth]{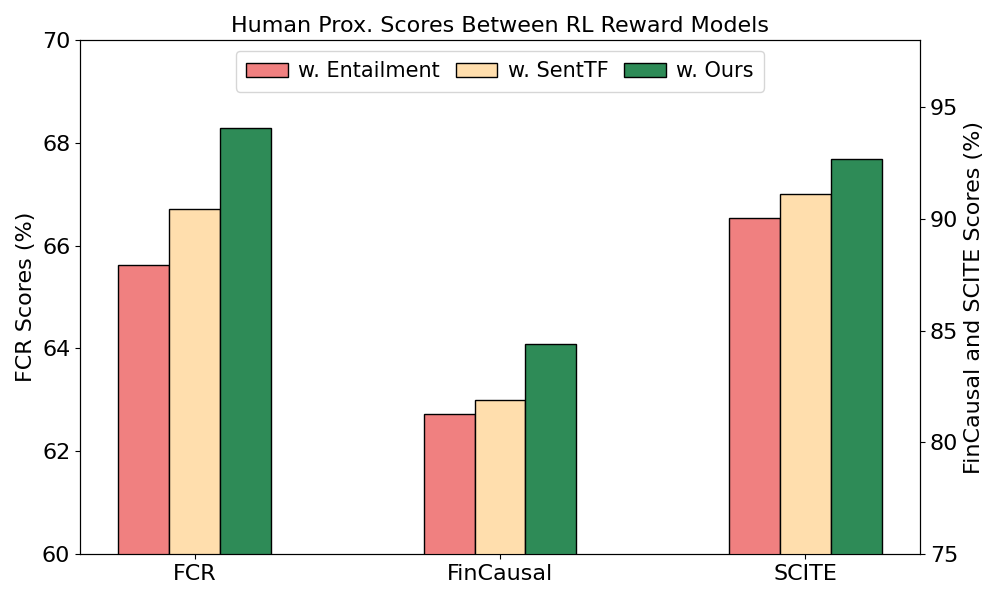}
    \end{subfigure}
    
    \caption{Ablation results for reward model with Human Proximal metric
    showing our reward model performs the best.}
    \label{fig:rl-ablation-human}
\end{figure}

To analyse the effects of our reward model trained on the human annotation
dataset, we replace it with two representative alternatives: an entailment-based
Natural Language Inference
model~\citep{williamsBroadCoverageChallengeCorpus2018} and SentenceTransformer
(SentTF). Entailment represents whether the model output is a logical
consequence of the input text, indicating the cause-effect relation.
SentenceTransformer is a pre-trained sentence embedding method, which we use to
embed the gold extraction and model outputs, with the score being their
normalised cosine similarity. Our reward model achieves the best Human Proximal
score across the three datasets (Figure~\ref{fig:rl-ablation-human}).

\paragraph{Tolerance to Wording Variance.} 

Our reward model trained on the human annotation data captures the high-level
semantic overlapping between gold extraction and model outputs. It is also
capable of identifying the correctness of model outputs through source text
understanding. Therefore, we use the "without EM (w/o EM)" metric to measure the
percentage of correctly generated samples that are not exactly matched with the
provided reference. This highlights the main improvement over automated metrics,
where we can recognise results that are correct but would have otherwise been
marked as incorrect because of their inexact result, showing clear advantages
for our evaluator over using Entailment or SentenceTransformer.

\subsection{Weak Supervision Evaluation}\label{sec:weak_evaluator}

The results in Table~\ref{tab:metric-correlation} show an evaluator model highly
aligned with human preference data. However, this requires a time-consuming and
expensive process of manual annotation. To decrease the reliance on this
process, we looked for ways to decrease the training set size.

\begin{figure}[ht]
    \centering
    \includegraphics[width=0.48\textwidth]{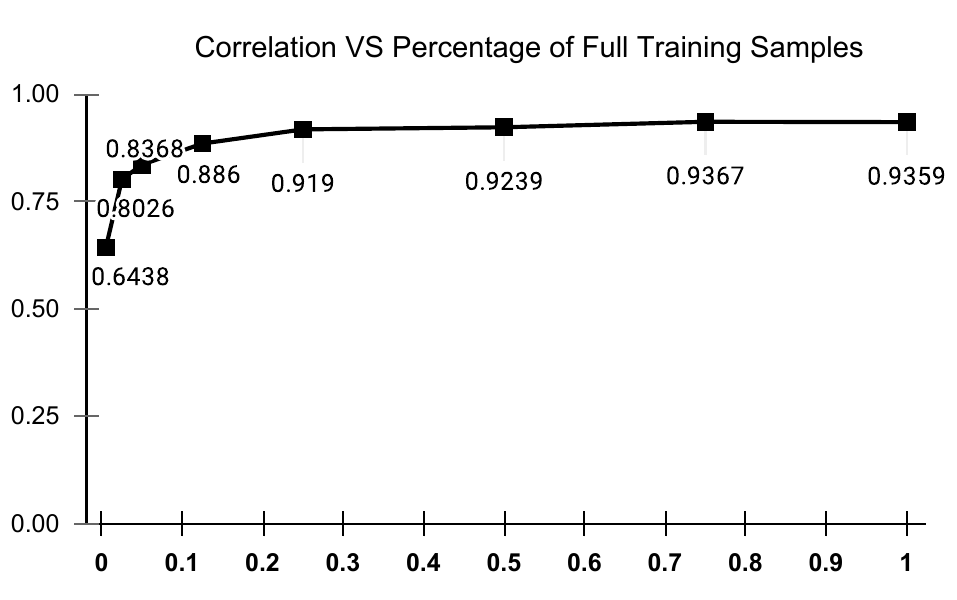}
    \caption{Evaluator agreement with human annotation by percentage of FCR data used.}
    \label{fig:agreement-by-split}
\end{figure}

We chose the FCR-based DeBERTa-Valid evaluator from
Section~\ref{sec:evaluation}, as it showed the highest agreement with human
evaluation and other datasets. We experimented with subsets of different sizes
and evaluated their performances. The results
(Figure~\ref{fig:agreement-by-split}) show we can decrease the training set size
with a small impact on the human agreement of the resulting evaluator. This
motivated us to pursue a way to train a high-quality evaluator with less data.

Our weak supervision process has three steps. First, we sample a random subset
of the training data as our initial supervised dataset and use it to train a
partial evaluator. Second, we apply this partial evaluator to the remaining
data, which we treat as unsupervised. We obtain the weak classification labels
and the confidence of the evaluator for each entry and use a filtering process
to determine which ones to keep. Third, we combine the filtered entries with the
original subset and train a final evaluator. Our filtering process separates the
weak labels into positive and negative sets, and for each set, takes the top
75\% entries by confidence, so the final filtered set has an equal number of
positive and negative entries.

Table~\ref{tab:weak-rl} shows the results of our weak supervision experiments.
We experimented with different subset sizes and found that the 50\% subset
achieves the best performance in terms of the Human Proximal and w/o EM metrics.
It also matches the performance of the Full RL model, showing we can
successfully decrease the reliance on human-annotated data without a performance
cost.




\begin{table}[ht]
    \centering
    \resizebox{\columnwidth}{!}{
    \begin{tabular}{lrrrrr}
        \toprule
        \textbf{Model} & \textbf{P} &\textbf{R} & \textbf{F1} & \textbf{HumanProx.} & \textbf{w/o EM $\uparrow$} \\ 
        \midrule

        \textsl{SFT}          & 80.02 & 80.48 & 80.96 & 64.42 & 10.29 \\
        \textsl{Full RL}      & \textbf{82.85} & 82.03 & 81.16 & 68.29 & 12.71 \\
        \textsl{30\% + weak}  & 80.28 & \textbf{84.19} & \textbf{82.18} & 68.37 & 12.63 \\
        \textsl{50\% + weak}  & 80.11 & 84.18 & 82.09 & \textbf{68.86} & \textbf{13.07} \\
        \textsl{80\% + weak}  & 81.18 & 82.23 & 81.72 & 67.41 & 11.60 \\

        \bottomrule
    \end{tabular}
    }
    \caption{RL with weakly-supervised models, showing the weakly-supervised
    variants are able to match the fully-trained model performance.}
    \label{tab:weak-rl}
\end{table}




\section{Conclusion}\label{conclusion}

We have explored several evaluation approaches to address the inherent ambiguity
of the causal event extraction task. We find that using a generative model to
perform extraction performs well, but that evaluation with automated metrics is
challenging. Our findings demonstrate the ability to faithfully reproduce human
evaluation results using a DeBERTa-based classifier trained on human evaluation
of extraction outputs. We also apply the evaluator as a reward model to
Reinforcement Learning, further aligning our generative extractor model to
human preferences. 

We explore multiple datasets, showing how our approach can be generalised and
employed our trained evaluator in a transfer setting, reducing the need for
further annotation of new data. Finally, we propose a weak-to-strong approach
where we only use a subset of annotated data to train a weakly-supervised
evaluator that can match the performance of the fully-trained version.

\section*{Limitations}\label{limitations}

The datasets we used are limited to ones where the causes and effects are spans
of the source text. Our approach does not work well with datasets where the
events are instead represented by trigger words, as is common in other datasets,
or when the answers are free text, not spans of the source text.

Another limitation is how we define the input of our evaluation. We require the
reference and without it, the evaluator does not perform well. This means we are
limited to datasets where we have such a reference, preventing us from applying the
evaluator to those with blind data where we only have the source text.

\section*{Acknowledgements}
This work was supported in part by the UK Engineering and Physical Sciences
Research Council (EPSRC) through a Turing AI Fellowship (grant no. EP/V020579/1,
EP/V020579/2) and a New Horizons grant (grant no. EP/X019063/1).

\bibliography{ref}
\clearpage
\newpage
\appendix
\setcounter{table}{0}
\renewcommand{\thetable}{A\arabic{table}}

\section{Dataset Transformation}

Our chosen datasets come in different formats, which we must transform into our
tagged format. FCR is a collection of JSON files, where each entry
contains the text and character indices for the cause and effect spans.
FinCausal contains semicolon-separated CSVs, where each entry contains the input
text and each cause effect spans as text. SCITE comprises XML files, where each
item is a tagged representation of the sentences and their spans. 

We convert them to a common format that is used as the base for all of our
models: a tagged representation, shown in Figure~\ref{fig:fincausal-example}.
For FinCausal and SCITE, which do not contain relations like FCR does, we
hard-code the Relation to `cause'. 

The original SCITE dataset has examples with more than one relation, which
our models do not support. We opted to use only the first causal relation
for each example.

\section{Further Dataset Statistics}

Table~\ref{tab:datasets-count} in the main text shows the count of instances per
dataset and split. We now show the average number of words for the source text,
cause and effect clauses in Table~\ref{tab:datasets-words}.

\begin{table}[htb]
    \centering
    \begin{tabular}{lccc}
    \toprule[1pt]
       \textbf{Dataset} & \multicolumn{3}{c}{\textbf{Average number of words}}  \\
       & Context & Cause & Effect \\
       \midrule
       FCR & 31.37 & 10.43 & 10.79 \\
       FinCausal & 42.77 & 18.23 & 17.20 \\
       SCITE & 18.68 & 2.15 & 2.03 \\
    \bottomrule[1pt]
    \end{tabular}
    \caption{Dataset statistics: average number of words per part.}
    \label{tab:datasets-words}
\end{table}

\section{Implementation Details}

We used the KL divergence during training to ensure that the policy does not
deviate too much from the format it learned during supervised fine-tuning (SFT).
We found that some of the batches during RL training would lead to very high KL
values, which would move the model too far in a given direction, often leading
to parameter collapses (i.e. model weights going to NaN or infinity) or
degenerate output (no longer recognisable as structured text).

To prevent this, we found that skipping batches with high KL values (over 2)
made training considerably more stable, as we only applied updates from batches
whose output was not too far from the reference model. The downside is that
this slows down training, as skipping batches means fewer updates, potentially
leaving the policy in a local optimum. In our experience, this trade-off was
worth it, considering we still achieved improvements in all our main RL
experiments.

\paragraph{Hyperparameters.} The SFT models used FLAN-T5-Large as the base. The
hyperparameters were the same across all datasets: input sequence length of 128
tokens, 20 training epochs, fixed learning rate of 0.0001 and greedy decoding
for generation. We used an early-stopping scheme with the patience of 5 epochs
without improvement based on the token F1 metrics.

The RL models were mostly similar, too: we used a single epoch, with the PPO
process using a learning rate of 0.00014. The initial KL coefficient varied by
dataset, with FCR using 0.4, SCITE using 0.2 and FinCausal 0.05. For generation,
the RL models used beam search (2 beams) with multinomial sampling. Other
parameters used the default values from the Transformers and TRL libraries.
Other configuration options, such as reward normalisation and scaling, did not
lead to any improvements. We found the RL models to be highly sensitive to
the hyperparameters.

The evaluation (reward) model was based on DeBERTa-V3-xsmall. Its input
sequence length was 400 tokens (to fit the input context and reference
extraction), learning rate of 0.00001 and 100 epochs, with early stopping
patience of 10 epochs without improvement based on the classification F1 score.
The reward models were largely robust across different hyperparameter values and
even sizes: with larger DeBERTa models not leading to significant improvements,
we preferred using the smallest model to decrease memory concerns when using it
alongside the larger FLAN-T5 model

\section{Software Used}

\paragraph{Versions.} We used
Transformers\footnote{\url{https://github.com/huggingface/transformers}} 4.33 to
train the FLAN-T5 and DeBERTa LLMs. For RL training, we used
TRL\footnote{\url{https://github.com/huggingface/trl}} 0.8.6. All experiments
were run using Python 3.12 on Ubuntu 20.04 with an NVIDIA A100 40 GB GPU running
CUDA 12.2. We also used NumPy\footnote{\url{https://numpy.org}} 1.24 and
PyTorch\footnote{\url{https://pytorch.org}} 2.0.

\paragraph{Licenses.} From the software mentioned above, NumPy and PyTorch use
the BSD license, TRL and Transformers use Apache-2.0, and Python uses the PSF
license. The original code for this project is licensed under GPL-3.0.

\paragraph{AI assistance.} GitHub
Copilot\footnote{\url{https://github.com/features/copilot}},
ChatGPT\footnote{\url{https://chat.openai.com/}} and
Claude\footnote{\url{https://claude.ai}} were used to assist in the development
of the code, while Perplexity\footnote{\url{https://perplexity.ai/}} was
used for general queries.

\section{Human Annotation}\label{sec:annotation-platform}

\begin{figure}[ht]
    \centering
    \includegraphics[width=0.48\textwidth]{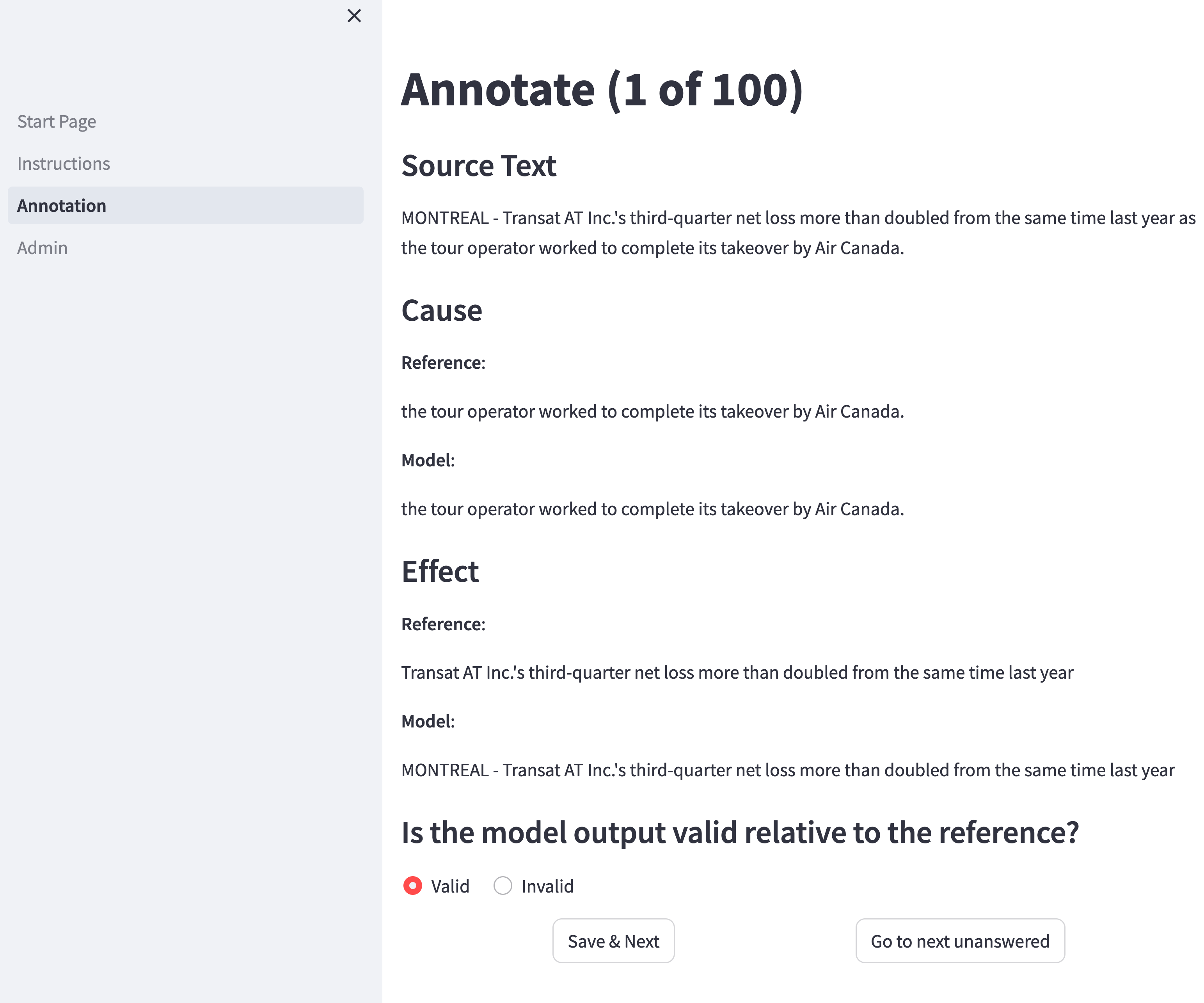}
    \caption{Screenshot of our annotation platform showing an example from the
    FinCausal dataset}
    \label{fig:annotation-platform}
\end{figure}

We built an online annotation platform using
Streamlit\footnote{\url{https://streamlit.io}} version 1.35. It was deployed on a
Digital Ocean\footnote{\url{https://www.digitalocean.com}} Droplet.
Figure~\ref{fig:annotation-platform} shows a screenshot of the annotation page
of the platform with an example from the FinCausal dataset.

The users were able to read the source text and compare the reference and model
outputs for each entry before selecting whether the entry was `valid' or
`invalid'. The platform saved the answers as soon as they were confirmed and
allowed the users to leave and return later to continue from where they stopped.

\end{document}